\documentclass[sigconf]{acmart}
\AtBeginDocument{%
  }

\setcopyright{acmlicensed}
\copyrightyear{2026}
\acmYear{2026}
\acmDOI{XXXXXXX.XXXXXXX}
\acmConference[LLM4Code'2026]{Make sure to enter the correct
  conference title from your rights confirmation email}{April
  2026}{Rio de Janeiro, Brazil}

\acmISBN{978-1-4503-XXXX-X/2018/06}

\usepackage{textcomp}
\usepackage{todonotes}
\usepackage{amsfonts}
\usepackage{soul}
\usepackage{algorithm} 
\usepackage{algorithmicx}
\usepackage{algpseudocode}
\usepackage{tablefootnote} 

\usepackage[T1]{fontenc}
\usepackage{todonotes}
\usepackage{booktabs}
\usepackage{multirow}
\usepackage{multicol}
\usepackage{array}
\usepackage{graphicx}
\usepackage{caption}
\usepackage{subcaption}
\usepackage{enumitem}
\usepackage{amsmath}
\usepackage{bbold}
\usepackage{xcolor}
\usepackage{colortbl}
\usepackage{pgfplots}
\usepackage{pgfplotstable}
\usepackage{tabularx}
\usepackage{mathtools}
\usepackage{collcell}
\usepackage{placeins}
\usepackage{pifont}
\usepackage{siunitx}
\usepackage{float}
\usepackage{adjustbox}
\usepackage{listings}
\usepackage{xcolor}
\usepackage{tcolorbox}

\usepackage{longtable}
\usepackage{makecell}

\newcolumntype{H}{c}  % Header column (for text, no coloring)
\newcolumntype{C}{>{\collectcell\vcellcolor}c<{\endcollectcell}} % Numeric column with coloring
\newcolumntype{X}{>{\raggedright\arraybackslash}p{1.5cm}} % Wrapped text column for first column
\newcolumntype{T}{>{\hsize=1.5\hsize\centering\arraybackslash}X} % Wider Total column
\newcolumntype{S}{>{\hsize=1\hsize\centering\arraybackslash}X}   % Standard numeric columns

\NewDocumentCommand\vcellcolorz{m}{%
  \centering
  \ifdim#1pt=\minvalz pt\cellcolor{tabminz}%
  \else% 
    \ifdim#1pt=\maxvalz pt\cellcolor{tabmaxz}%
    \else%
      \ifdim#1pt<0.5pt
        \cellcolor{lowerz!![\fpeval{round(\gradstepsz*(#1-\minvalz)/0.5)}]}%
      \else
        \cellcolor{higherz!![\fpeval{round(\gradstepsz*(#1-0.5)/0.5)}]}%
      \fi
    \fi%
  \fi
  \num[round-mode=places,round-precision=1,mode=text,detect-all]{\fpeval{#1*100}}\%
}

\NewDocumentCommand\vcellcolorredz{m}{%
  \centering
  \ifdim#1pt=\minvalz pt\cellcolor{tabminz}%
  \else
    \ifdim#1pt=\maxvalz pt\cellcolor{tabmaxz}%
    \else
      \ifdim#1pt<0.5pt
        \cellcolor{lowerz!![\fpeval{round(\gradstepsz*(#1-\minvalz)/0.5)}]}%
      \else
        \cellcolor{higherz!![\fpeval{round(\gradstepsz*(#1-0.5)/0.5)}]}%
      \fi
    \fi
  \fi
  \textcolor{red}{\num[round-mode=places,round-precision=1,mode=text,detect-all]{\fpeval{#1*100}}\%}
}

\newcolumntype{K}{c}  
\newcolumntype{F}{>{\collectcell\vcellcolorz}c<{\endcollectcell}} % Numeric column with coloring using vcellcolorz
\newcolumntype{L}{>{\raggedright\arraybackslash}p{1.5cm}} % Wrapped text column
\newcolumntype{M}{>{\hsize=1.5\hsize\centering\arraybackslash}L} % Wider Total column
\newcolumntype{N}{>{\hsize=1\hsize\centering\arraybackslash}L}   % Standard numeric columns

\newcommand\maxval{1}     % Max value 
\newcommand\minval{0}     % Min value 
\newcommand\gradsteps{10} % Number of gradient steps

\colorlet{tabmin}{red!80}   % Worst value color (red)
\colorlet{tabmid}{yellow!40} % Mid value color (yellow)
\colorlet{tabmax}{green!70}  % Best value color (green)

\definecolorseries{lower}{rgb}{last}{tabmin}{tabmid}
\definecolorseries{higher}{rgb}{last}{tabmid}{tabmax}
\resetcolorseries[\gradsteps]{lower}
\resetcolorseries[\gradsteps]{higher}

\NewDocumentCommand\vcellcolor{m}{%
  \centering
  \ifdim#1pt=\minval pt\cellcolor{tabmin}%
  \else% 
    \ifdim#1pt=\maxval pt\cellcolor{tabmax}%
    \else%
      \ifdim#1pt<0.5pt
        \cellcolor{lower!![\fpeval{round(\gradsteps*(#1-\minval)/0.5)}]}%
      \else
        \cellcolor{higher!![\fpeval{round(\gradsteps*(#1-0.5)/0.5)}]}%
      \fi
    \fi%
  \fi
  \num[round-mode=places,round-precision=1,mode=text,detect-all]{\fpeval{#1*100}}\%
}

\NewDocumentCommand\vcellcolorred{m}{%
  \centering
  \ifdim#1pt=\minval pt\cellcolor{tabmin}%
  \else
    \ifdim#1pt=\maxval pt\cellcolor{tabmax}%
    \else
      \ifdim#1pt<0.5pt
        \cellcolor{lower!![\fpeval{round(\gradsteps*(#1-\minval)/0.5)}]}%
      \else
        \cellcolor{higher!![\fpeval{round(\gradsteps*(#1-0.5)/0.5)}]}%
      \fi
    \fi
  \fi
  \textcolor{red}{\num[round-mode=places,round-precision=1,mode=text,detect-all]{\fpeval{#1*100}}\%}
}

\usepackage[T1]{fontenc}

\usepackage{listings}
\usepackage{xcolor}

\definecolor{darkgreen}{rgb}{0.0, 0.5, 0.0}
\definecolor{jsonBg}{RGB}{250, 250, 247}       % very light ivory
\definecolor{jsonString}{RGB}{20, 20, 20}    % medium gray   
\definecolor{jsonNumber}{RGB}{0, 96, 160}    % violet
\definecolor{jsonBool}{RGB}{0, 96, 160}        % blue
\definecolor{jsonNull}{RGB}{160, 0, 0}         % dark red
\definecolor{jsonPunct}{RGB}{120, 120, 120}    % medium gray
\definecolor{jsonComment}{RGB}{128,128,128}    % gray for comments if any

\lstdefinestyle{llmstyle}{
    backgroundcolor=\color{white},   
    basicstyle=\ttfamily\footnotesize,   
    frame=single,                
    rulecolor=\color{black},    
    breaklines=true,             
    keepspaces=true,             
    showstringspaces=false,      
    tabsize=4
}

\begin{document}

\title{An Initial Exploration of Contrastive Prompt Tuning to Generate Energy-Efficient Code}
%\title{Watts in the Code: Contrastive Prompt Tuning to Generate Energy-Efficient Software}

\author{Sophie Weidmann}
\email{sopwdm@gmail.com}
\affiliation{%
  \institution{University of Twente}
  \city{Enschede}
  \country{The Netherlands}
}

\author{Fernando Castor}
\email{f.castor@utwente.nl}
\affiliation{% 
  \institution{University of Twente}
  \city{Enschede}
  \country{The Netherlands}
}

\renewcommand{\shortauthors}{Weidmann and Castor}
\newcommand\gls[1]{#1}
\newcommand\glspl[1]{#1}
\newcommand\Gls[1]{#1}
\newcommand\Glspl[1]{#1}
\newcommand\glsentryname[1]{#1}

\begin{abstract}
Although LLMs are capable of generating functionally correct code, they also tend to produce less energy-efficient code in comparison to human-written solutions. As these inefficiencies lead to higher computational overhead, they are in direct conflict with Green Software Development (GSD) efforts, which aim to reduce the energy consumption of code. To support these efforts, this study aims to investigate whether and how LLMs can be optimized to promote the generation of energy-efficient code. To this end, we employ Contrastive Prompt Tuning (CPT). CPT combines Contrastive Learning techniques, which help the model to distinguish between efficient and inefficient code, and Prompt Tuning, a Parameter-Efficient Fine Tuning (PEFT) approach that requires only a fraction of the cost of traditional fine tuning. This study evaluates CPT on Python, Java and C++ coding problems across three different models to provide a comprehensive evaluation. The method achieves consistent improvements in code accuracy for two models but efficiency gains vary by model, language and task complexity, indicating that improvements are not uniformly reliable.
\end{abstract}



\maketitle

\section{Introduction}

%Large language models (LLMs) have progressed rapidly from natural language processing problems to complex software engineering tasks. This resulted in a fundamental shift of how code is written, reviewed, and maintained \cite{zheng_survey_2024}. Code-centric models such as Codex \cite{chen_evaluating_2021}, CodeBERT \cite{feng_codebert_2020}, and CodeT5 \cite{wang_codet5_2021} were trained on massive, high-quality code corpora to perform especially well on code-related tasks. This progress has also driven the motivation to embed LLMs into every phase of the software development process. Tools like GitHub's Copilot are widely used in the industry and support developers in various programming tasks, especially code generation. According to a large-scale analysis of 934,533 GitHub Copilot users, developers accept nearly 30\% of Copilot’s code suggestions on average, with acceptance rates increasing over time as users become more familiar with the tool \cite{dohmke_sea_2023}.

The quality of LLM-generated code has become a subject of growing interest. LLMs can generate functionally-correct code in many cases~\cite{chen_evaluating_2021}. However, this is impacted by factors such as the programming language and the complexity of the task~\cite{Liu:2023:IYC,solovyeva_ai-powered_2025}. Furthermore, LLM-generated code has been found to be less efficient than human-written code~\cite{Erhabor:2025:MRP, solovyeva_ai-powered_2025, vartziotis_learn_2024}. This inefficiency creates a challenge for green software development efforts, as poorly optimized code can lead to increased computational costs and therefore higher energy consumption and carbon emissions.

An important point to consider is the trade-off that comes from using LLMs to create (optimized) code. The energy cost of inference for a model are high and might very well outweigh any savings introduced from generating more energy-efficient code. This is especially of concern as newer models were found to improve the quality of the code at the expense of longer inference times \cite{openai_openai_nodate, zeff_openais_2024}. %This naturally raises the question of whether the goal of this work makes sense at all, also in regards to green software development efforts.
We believe that, since LLMs are already widely used for code generation nonetheless \cite{dohmke_sea_2023}, it is still of value to ensure that they default to generating efficient rather than inefficient code. In addition, the cloud infrastructures where many models are typically executed tend to use energy more efficiently than developer workstations and local IT services within companies and universities~\cite{Patterson:2022:CFM}. This factor, combined with the need to run code on battery-powered devices provides additional motivation to investigate the ability of LLMs to generate energy-efficient code. 

% There are many approaches to improve model performance under different scenarios models, e.g., fine-tuning, prompt engineering, prompt-tuning, etc. Why are they not great to improve the efficiency of the generated code? This is the essential question to be answered to create a nice connection to contrastive prompt tuning and contrastive learning (for actual fine-tuning). 
In this paper, we present preliminary, exploratory work investigating a potential path to make LLM-generated code more energy-efficient: contrastive prompt tuning (CPT). Prompt tuning~\cite{lester_power_2021} leverages gradient descent to train small sets of parameters that are added to the model, also known as ``soft prompts''. The latter are appended to the prompt provided to the LLM to improve the results they produce. The soft prompt is iteratively refined similarly to a fine-tuning process, but the base model is kept frozen, i.e., its parameters do not change. Peng et al.~\cite{peng_model_2024} show that prompt tuning can outperform both traditional fine tuning and manual prompts while requiring significantly less resources than the former. We employ a contrastive approach to prompt tuning: soft prompts are optimized in terms of the difference between desired and undesired outputs and not how far predictions are from expected results. To the best of our knowledge, this is the first paper to explore contrastive approaches to improve quality aspects of generated code. 

We tune the prompts using pairs of functionally-equivalent programs where one is more energy-hungry and the other is more energy-efficient and employ these models to generate code in Python, C\texttt{++}, and Java. We built a dataset of such pairs combining programs from the Generation of Efficient Programs\footnote{https://github.com/CodeGeneration2/GEC-Dataset} (Python) and CodeNet\footnote{https://github.com/IBM/Project\_CodeNet} datasets (C++), as well as a dataset of solutions to programming problems from LeetCode  in Python and C++~\cite{solovyeva_ai-powered_2025}.

%----------------------------------------------------------------------------------------------------
\section{Background}

%In this section we briefly introduce the concepts of prompt tuning and contrastive learning.
%\vspace{0.3cm}

%----------------------------------------------------------------------------------------------------
\noindent\textbf{Prompt Tuning}~\cite{lester_power_2021} is a method to efficiently adapt language models to downstream tasks. As a parameter-efficient fine tuning~\cite{Liu:2022:FSP} method, prompt tuning optimizes only a small set of embeddings, the soft prompt, while keeping the base model frozen. By keeping the model's weights unchanged, its inherent behavior and acquired knowledge are preserved. This approach reduces computational overhead. The training process of a soft prompt closely resembles traditional fine-tuning.  
However, instead of updating the entire model, only parameters of the soft prompt are adjusted. This process is repeated iteratively, gradually refining the soft prompt to guide the model’s processing in a way that minimizes the loss.

\begin{figure}[tb]
    \centering
    \includegraphics[width=0.35\textwidth]{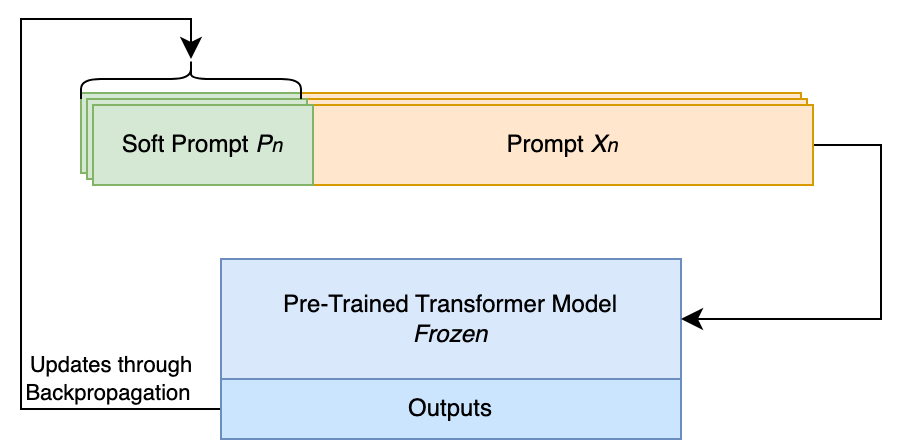}
    \caption{In prompt tuning, the soft prompt is optimized through backpropagation and the model remains frozen. Input consists of learnable soft prompt and  standard prompt.}
    \label{fig:prompt_tuning_embeddings}
    \vspace*{-0.3cm}
\end{figure}

Soft prompts are appended as additional tokens to the model's input embedding, and only the weight of these tokens are trained, the process being depicted in Figure \ref{fig:prompt_tuning_embeddings}. These tokens can be initialized in various ways, most commonly through random initialization, although heuristic-based approaches can also be used. Once initialized, the soft prompt is placed at the beginning of the input. Notably, this part of the prompt cannot be translated to natural language.
When the model processes the combined input, it considers both the soft prompt and the original input data. 
\vspace{0.3cm}

%----------------------------------------------------------------------------------------------------
\noindent\textbf{Contrastive learning} is a self-supervised or supervised learning approach that aims to learn useful representations by pulling similar instances closer while pushing dissimilar instances apart in the embedding space~\cite{hadsell_dimensionality_2006}. It has been used in various ML tasks, especially representation learning for vision and NLP tasks~\cite{le-khac_contrastive_2020}. 

When training a model using a contrastive learning approach, each batch contains at least one positive sample \( x^+ \), and one or more negative samples \( x^- \). Some contrastive learning techniques also require an anchor sample \( x \), which serves as a reference point to compare the positive \( x^+ \) and negative \( x^- \) samples against. In that case, the positive sample \( x^+ \) is a semantically similar instance to the anchor, while the negative sample \( x^- \) is a dissimilar instance from the dataset. 
The embeddings produced from the input samples during training exist in a high-dimensional embedding space, where distances reflect similarity: a smaller distance indicates higher similarity, while a larger distance corresponds to lower similarity. Different similarity or distance metrics can be used to measure the distance, such as cosine similarity or euclidean distance. Once these pairwise similarities are computed, a contrastive loss function is used to adjust the embedding so that similar samples are pulled closer together and dissimilar samples are pushed further apart in the embedding space.%, as illustrated in Figure~\ref{fig:contrastive_learning}.
Different types of contrastive loss functions exist, which vary in how they weigh or structure these relationships. Examples include Pair Loss, InfoNCE, and Loss Triplet.

In this work, we leverage a contrastive approach to perform prompt tuning, instead of training the model. Prompt tuning involves adjusting the embeddings that are appended to the prompt. We use a contrastive loss function to guide these adjustments.

%----------------------------------------------------------------------------------------------------
\section{Related Work}

Solovyeva et al.~\cite{solovyeva_ai-powered_2025} measured the energy use of solutions to hard LeetCode problems written in Java, Python, and C\texttt{++} and generated by four frontier models. They found out that good solutions from LeetCode consistently outperform generated solutions. More recently Apsan and colleagues~\cite{Apsan:2026:GEE} investigated the effectiveness of using different prompting strategies and found out that code written by a human specialist consistently outperforms LLMs generating Python code. A different approach is taken by Du et al.~\cite{Du:2025:ARL}, who proposed a system named Afterburner that employs reinforcement learning to improve the efficiency of generated programs. Unlike previous work, we investigate the potential of CPT to make pre-trained LLMs to generate more energy-efficient code. 

Contrastive learning approaches have been explored in multiple scenarios, particularly to improve the discriminating ability of computer vision models~\cite{Chen:2020:SFC}. Usually, these approaches involve training or performing full fine-tuning of a model using a paired dataset and a contrastive loss function. Contrastive learning has also been used in coding-related tasks, such as code generation~\cite{jain_contraclm_2023,Zhang:2024:CRT}, search and classification~\cite{Bui:2021:SSC, Xu:2025:DLG}, summarization~\cite{Bui:2021:SSC}, ranking~\cite{jain_contraclm_2023}, and even to help distinguish between human-written and ChatGPT-generated code~\cite{Xu:2025:DLG}.
All these papers highlight a common trend in the use of contrastive learning approaches: they involve teaching a model to distinguish (or associate) pairs of input elements. This ability to discriminate can then be leveraged to help the model perform some useful task. To the best of our knowledge, previous work has not attempted to improve runtime characteristics of generated or preexisting code, focusing instead of their structural and semantic aspects. In this work we attempt to exploit differences in the structure and constructs employed by functionally-equivalent programs that exhibit different energy usage to help language  models produce more efficient code. Furthermore, none of the aforementioned approaches employs prompt tuning; they require full-fledged fine-tuning, which is computationally and environmentally expensive. 

\section{Methodology}\label{sec:eval}

% The main question that this work tries to answer is:

% \begin{quote}
%     \emph{\textbf{Main RQ}: To what extent can Contrastive Prompt Tuning improve the energy efficiency of LLM-generated code while preserving functional correctness?}
% \end{quote}
% To ensure a thorough investigation of this topic, the main research question has been divided into the following sub-questions. 

In this work we investigate how contrastive prompt tuning can improve the energy efficiency of LLM-generated code while preserving its functional correctness. Below we introduce the three research questions we tackle. 

\vspace{0.1cm}
\noindent
\textbf{RQ1}: \textit{To what extent do different contrastive loss functions influence the effectiveness of CPT to generate correct and efficient code?}
%Contrastive learning allows models to learn the difference between desirable and undesirable outcomes. The learning process is informed by the employed contrastive loss function. In this work, we examine two contrastive loss functions, InfoNCE and Triplet Loss. We investigate both the correctness and efficiency of the generated code.

\vspace{0.1cm}
\noindent
\textbf{RQ2}: \textit{How do improvements from CPT vary across model types?}
%Models that are pre-trained on domain-specific data often exhibit a foundational understanding that is beneficial for domain-specific tasks. In the context of code generation, LLMs that were pre-trained on extensive code corpora may perform better in producing efficient and optimized solutions. This work applied CPT separately to three different models trained in different manners by their authors (general-purpose, code-tuned, and code-first) to evaluate if and how CPT affects each model's ability to generate correct and energy-efficient code. 
%, and to determine if results generalize across model types. 
 
\vspace{0.1cm}
\noindent
\textbf{RQ3}: \textit{How do improvements from CPT vary across PLs}
%Programming languages differ significantly in syntax and performance. These differences may influence how LLMs interpret, generate, and optimize code. For example, Python’s dynamic nature might present different challenges compared to the static nature and memory management requirements of C\texttt{++}. 
%Similarly, the potential for energy-efficiency improvements may differ between simple and more complex tasks. 
%This work evaluated CPT across three languages, Python, Java, and C++, examining whether CPT’s benefits generalize or require language-specific adaptations.

\vspace{0.1cm}

\begin{figure}[tb]
    \centering
    \includegraphics[width=0.45\textwidth]{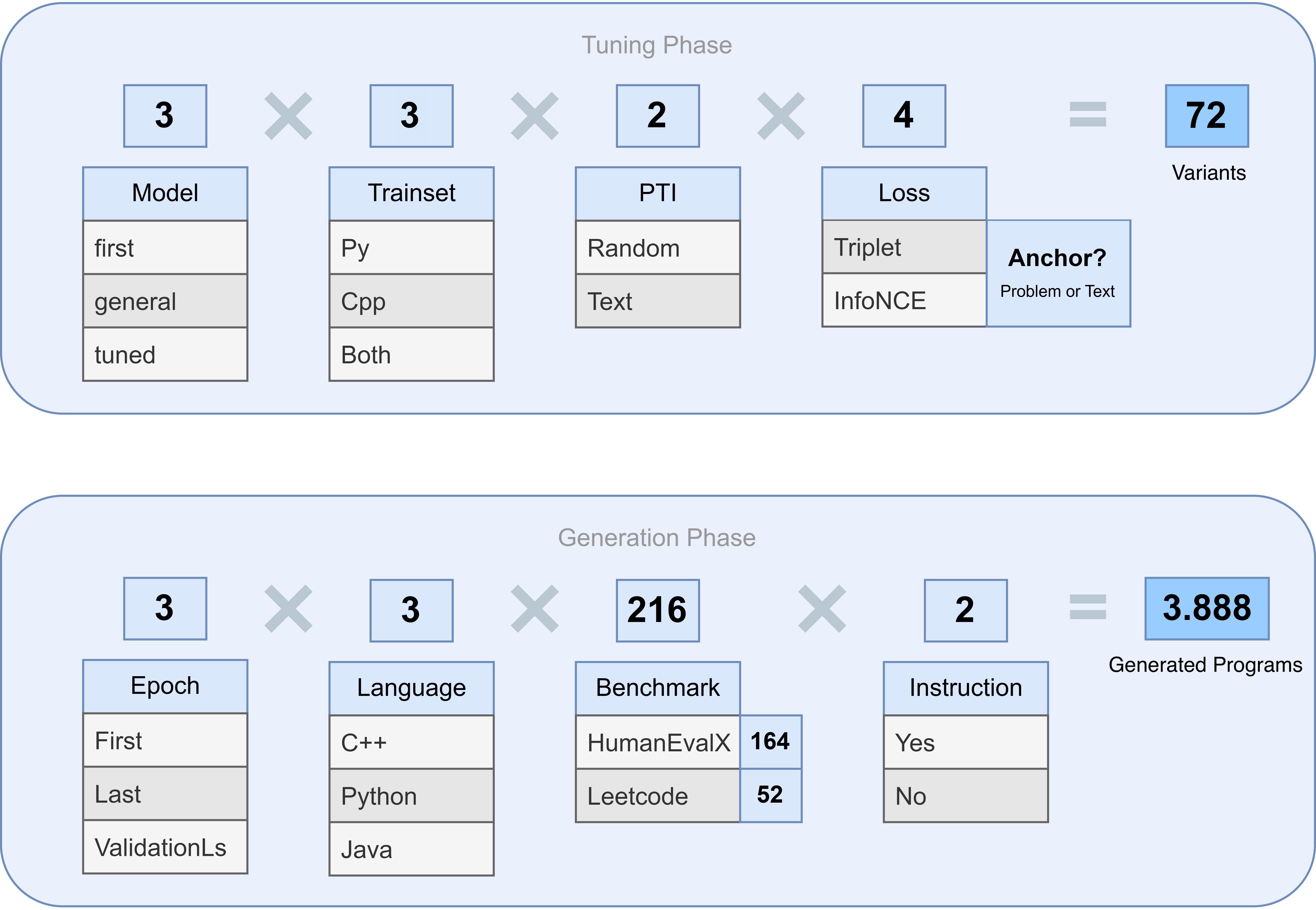}
    \caption{Overview of the experimental design and configuration space for the training and generation phases. }
    \label{fig:exp_config_space}
\end{figure}

Figure~\ref{fig:exp_config_space} gives an overview of the experimental design and configuration space of the study. It shows two phases: training and generation. After the latter, the generated programs are tested for correctness and executed and measured for performance. The following sections explain the elements of this figure.

%%%%%%%%%%%%%%%%%%%%%%%%%%%%%%%%%%%%%%%%%%%%%%%%%%%%%%%%%%%%%%%%%%%%%%

\subsection{Model and Dataset Selection}
\label{subsec:model_selection}

%The language  models in this work are supported by the the Hugging Face Transformers library\footnote{Documentation: \url{https://huggingface.co/docs/transformers}}. 

\noindent
\textbf{Model selection.} To assess the effectiveness of CPT on energy-efficient code generation, this work compared three models. They differ in their architectural approach, by being built or tuned for code or not. Table~\ref{tab:selected_models} lists the three studied models.  

\begin{table}[bt]
  \begin{scriptsize}
    \centering
    \renewcommand{\arraystretch}{1.2}
    \begin{tabular}{|l|l|l|c|}
        \hline
        \textbf{Abbrevation}  & \textbf{Model Type} & \textbf{Model Name} & \textbf{Parameters} \\
        \hline
        \textbf{\texttt{general}} & General-Purpose & Phi-3 Mini\tablefootnote{\url{https://huggingface.co/microsoft/Phi-3-mini-4k-instruct}} & $\sim$ 3.8B \\
        \hline
        \textbf{\texttt{tuned}} &  Code-Tuned & CodeLlama-7B\tablefootnote{\url{https://huggingface.co/codellama/CodeLlama-7b-hf}} & $\sim$ 7B \\
        \hline
        \textbf{\texttt{first}} & Code-First & DeepSeek Coder\tablefootnote{\url{https://huggingface.co/deepseek-ai/deepseek-coder-7b-instruct}} & $\sim$ 7B \\
        \hline
    \end{tabular}
    \caption{Selected models for the experiment and evaluation.}
    \label{tab:selected_models}
  \end{scriptsize}
\end{table}

\vspace{0.1cm}
\noindent
\textbf{ Training datasets.} 
This work includes two distinct dataset types: Training dataset and Benchmark dataset (Table~\ref{tab:dataset_overview}). 
%To improve the effectiveness of LLMs in generating energy-efficient code, 
CPT was applied using efficient and inefficient code pairs in Python and C\texttt{++} to train the three models. The data for creating these code pairs was taken from two publicly-available datasets. These datasets include a wide range of programming problems and solutions, and detailed information per problem, such as the execution time and memory usage, which were used to determine which solutions ran efficiently and which ones did not. 
% An example entry can be found in Appendix \ref{app:trainset_example}. 
Throughout the paper, we refer to three training datasets, \textbf{Python}, \textbf{C\texttt{++}}, and \textbf{Both}. Each entry in these datasets includes one efficient and several inefficient solutions. The first two datasets consist of programs written in Python and C++, respectively. For \textbf{Python}, the pairs of programs were taken from the \textit{Generation of Efficient Code (GEC)} dataset\footnote{\url{https://github.com/CodeGeneration2/GEC-Dataset}}.
%, which consists of 3,712 coding problems and 31,577 efficient-inefficient code pairs of Python programming tasks. 
The \textbf{C\texttt{++}} pairs come from the  \mbox{\textit{CodeNet}} dataset\footnote{\url{https://github.com/IBM/Project_CodeNet}} \cite{puri_codenet_2021} dataset of coding problems. 
%CodeNet was designed for program understanding and generation. It contains millions of code snippets across various programming languages. The dataset includes 13,916,868 submissions covering 4,053 problems, with 57\% of all submissions written in C\texttt{++}. 
Finally, the \textbf{Both} dataset consists of the union of the \textbf{Python} and \textbf{C\texttt{++}} datasets being used to train the models.

\begin{table}[tb]
    \centering
    \renewcommand{\arraystretch}{1.2}
    \scriptsize
    \begin{tabular}{|l|c|c|}
        \hline
        \multicolumn{3}{|c|}{\textbf{Training Sets}} \\
        \hline
        \textbf{Dataset} & \textbf{Language(s)} & \textbf{Size} \\
        \hline
        Python & Python & 3,548 entries \\
        \hline
        C\texttt{++} & C\texttt{++} & 3,921 entries \\
        \hline
        Both & Python, C\texttt{++} & 7,469 entries \\
        \hline\hline
        \multicolumn{3}{|c|}{\textbf{Benchmark Sets}} \\
        \hline
        \textbf{Dataset} & \textbf{Language(s)} & \textbf{Size} \\
        \hline
        Leetcode & Python, Java, C\texttt{++} & 52 problems per language \\
        \hline
        HumanEval-X & Python, Java, C\texttt{++} & 164 problems per language \\
        \hline
    \end{tabular}
    \caption{Overview of training and benchmark datasets used in this work.}
    \label{tab:dataset_overview}
    \vspace*{-0.6cm}
\end{table}

\vspace{0.1cm} 
\noindent
\textbf{Benchmark datasets.}  
\label{subsec:workload}
A diverse set of programming problems was used to ensure a comprehensive evaluation, taken from two code-generation benchmarks, HumanEval-X and LeetCode. %An example of a benchmark entry can be found in Appendix \ref{app:benchmark_example}. 
\textbf{HumanEval-X}~\cite{zheng_codegeex_2023}
%To measure the energy-efficiency for a wide variety of coding problems, the \textit{HumanEval-X} benchmark \cite{zheng_codegeex_2023} was used. This benchmark 
covers five programming languages with 164 simpler programming problems. For this work, only the HumanEval-X benchmarks for C\texttt{++}, Java, and Python were used. The \textbf{LeetCode} dataset~\cite{solovyeva_ai-powered_2025} consists of ``hard'' problems and solutions to them taken from LeetCode. It contains 52 coding problems across C\texttt{++}, Java, and Python.

%------------------------------------------------------------------------------------------------------------
\subsection{Contrastive Prompt Tuning}
CPT was implemented using the PEFT library from HuggingFace\footnote{Documentation: \url{https://huggingface.co/docs/peft}}. This library offers an implementation of prompt tuning where a set of tokens is introduced and optimized, while the base model weights remain frozen. 
%A thorough explanation of \gls{prompttuning} can be found in Section \ref{sec:prompttuning}. For \Gls{prompttuning} initialization, both random and text-based strategies were evaluated, with a default prompt length of 50 tokens. 
The default setup includes a learning rate of $5\cdot10^{-5}$. 
During training (prompt tuning), the code pairs of the trainsets are tokenized and processed by the model's embedding layer. This results in embedding for efficient and inefficient code snippets, which serve as the foundation for contrastive learning. The soft prompt was initialized as a text string (\textit{``Generate energy-efficient code.''}) since  preliminary experiments with random initialization resulted in very low accuracy.

In this work, we examined two contrastive loss functions~\cite{Khosla:2020:SCL}, Triplet Loss and InfoNCE. For the former, we employed the implementation available in the PyTorch Metric Learning library\footnote{Documentation: \url{https://kevinmusgrave.github.io/pytorch-metric-learning/}}. The latter had to be manually implemented, as no suitable pre-built variant we could find supported the required structure comprising an anchor (similar to Triplet Loss), a corresponding positive (efficient solution), and multiple negatives (inefficient solutions), all tied to the same problem. 

%The contrastive loss functions adjust the embedding, thereby pulling efficient solutions closer together, while pushing inefficient ones farther apart in the embedding space. This step does not involve any text (or code) generation from the model. Instead, it focuses solely on learning \gls{embedding} representations that align efficiency with specific problem structures. 
%The batch size was initially set to 16. However, the code-tuned and code-first model, in combination with InfoNCE, required a batch size of 4 to run and to balance memory efficiency. This batch size is relatively small, but as analyzed by Xu et al. (2023) \cite{xu_making_2023} a small batch size is sufficient in a \gls{contrastivelearning} setting. 
%During each training step, backpropagation  is performed to iteratively update the soft prompt.

The validation cycle in this setup was limited due to the nature of contrastive learning and prompt tuning. Since the model is not generating an output during the training phase, common standard validation techniques are not applicable. Instead, any validation relied on techniques such as embedding separation metrics. These provide an indirect assessment of how well the model differentiates between efficient and inefficient solutions. This work used cosine similarity measurements, which track how close efficient embeddings are relative to inefficient ones. While improvements in the \gls{embedding space} can imply that the model is learning efficiency patterns, this is not a given. Therefore, efficiency improvements can only be fully validated post-training and post-generation. 
%This requires a separate evaluation phase where the generated code is executed and measured, as described in Section \ref{sec:evaluation_setup}. 

%However, before the phase of generating code, the most suitable training checkpoint (epoch) had to be selected for generation. In early experiments the checkpoint was based on the best validation loss reported. In later experiments, also the lowest epoch (epoch 1) and the highest epoch (epoch 20 or epoch 100) were used to examine the impact of minimal vs heavy training. 

%------------------------------------------------------------------------------------------------------------
\subsection{Code Generation and Execution}
\label{subsec:code_generation}

The quality of the generated code can be heavily influenced by the structure of the input prompt. To maintain the consistency across different programming languages and models, regardless of whether and how CPT was applied, the prompts in this work followed a standardized format. Each prompt had three components. The first was an optional \textbf{energy-efficient instruction}. Later experiments showed that the prompt-tuned models for Triplet and InfoNCE loss greatly profited from adding an instruction as part of the prompt. This instruction was meant to steer the model towards more energy-efficient code: \textit{"Write efficient code for the following problem:"}. The second was a \textbf{problem statement} in natural language, a description of the problem, the expected functionality of the solution, and input and output examples. Finally, the prompt also included a \textbf{function signature} in the respective programming language, specifying the function name and parameter types. This ensured that the generated solution matched a predefined structure and was syntactically correct. It also eased the process of comparing solutions for the same problem.

%The baseline and prompt tuned models were evaluated by using programming problems of varying complexity, as described in Section \ref{subsec:workload}. 
Each model was prompted using the described format. Key model configuration parameters were standardized. The maximum number of new tokens generated was capped at 512, and decoding was performed deterministically with the \texttt{do\_sample=False} parameter, which disables randomness in decoding, making the model always select the most likely next token. In this setup, the \texttt{temperature} parameter has no effect.%, as it only influences the sampling distribution when \texttt{do\_sample=True}. 

Before evaluating the generated outputs, the generated code solutions underwent formatting and cleaning. Some models generated extra functions or explanatory code which had to be removed. Therefore, more solutions could be evaluated by applying some post-generation steps and not just discarding them. However, the goal of this step was only to minimize errors caused by the aforementioned issues and not to fix incorrect code.

After processing the generated code, it was compiled (Java and C\texttt{++} only) and then executed to determine whether the test cases provided by the benchmark passed without raising errors. To evaluate the efficiency of the code, test suites were extended to run for 5-10 seconds, as this improves measurement accuracy. Any solutions that failed were excluded from further evaluation.  We collect execution data for three dependent variables: (i) accuracy, as characterized by pass\@1 code correctness, (ii) execution time (in ms), and (iii) energy consumption (in joules). For the first variable, higher is better and for the latter two, lower is better.

%-----------------------------------------------------------------------
\subsection{Environment}
%A controlled measurement environment is important to ensure the same conditions when evaluating the performance of the generated code. 
All solutions were executed on one machine with the same hardware and software configurations, a \textbf{MacBook Pro (15-inch, 2016)}. The hardware specifications are detailed in Table~\ref{tab:hardware_setup}. The generated code was executed under controlled settings and measured for CPU power consumption. C\texttt{++} solutions were compiled with Apple Clang 14.0.0 using the \texttt{-std=c++17} standard flag and the \texttt{-O0} optimization flag. Java solutions were compiled with \texttt{javac} and run with \texttt{java -ea -cp}. Python solutions were executed using \texttt{python3.12}. 
All code executions were conducted via the MacBook's integrated terminal application. The device was fully charged and connected to a power source. Lastly, to minimize any external influences, Wi-Fi and Bluetooth were disabled, and no additional applications or background processes were running during the execution.

\begin{table}[tb]
  \begin{scriptsize}
    \centering
    \renewcommand{\arraystretch}{1.2}
    \begin{tabular}{|l|l|}
        \hline
        \textbf{Component} & \textbf{Specification} \\
        \hline
        \textbf{Device} & MacBook Pro (15-inch, 2016) \\
        \hline
        \textbf{Processor} & 2.7 GHz Quad-Core Intel Core i7 \\
        \hline
        \textbf{Memory} & 16 GB 2133 MHz LPDDR3 \\
        \hline
        \textbf{Graphics} & Intel HD Graphics 530 (1536 MB) \\
        \hline
        \textbf{Storage} & 512 GB SSD \\
        \hline
        \textbf{OS} & macOS Montery v12.7.6 \\
        \hline
    \end{tabular}
    \caption{Hardware setup used for evaluation.}
    \label{tab:hardware_setup}
\end{scriptsize}
\vspace*{-0.5cm}
\end{table}

%-----------------------------------------------------------------------
\subsection{Measurement and Analysis}
To collect data about energy consumption and the runtime performance, we employed \textsf{powermetrics}, a built-in macOS system tool.  
Since it collects the power data at regular intervals and runs in its own process, the start and end of its execution had to be synchronized with the code execution process to ensure accurate measurements. The power and execution data were sampled at a frequency of 100 samples per second (100 Hertz). This makes it possible to capture fine-grained variations during the code executions

Each generated code solution was executed under identical workload conditions to maintain consistency. Every execution had a minimum duration of 5 seconds to allow stable energy readings. A cooldown period of 5 seconds was applied between consecutive runs. This mitigated the risk of thermal throttling affecting the results and avoids capturing tail energy between the trials. To account for minor fluctuations in the measurements of one problem, each solution was executed 10 times. The average value of the runs was used for further evaluation.

Outliers were removed using the Median Absolute Deviation~\cite{Leys:2013:DOS} method to improve robustness against extreme values. Given the small sample sizes and assumed non-normal distribution, the Mann-Whitney~U test was employed to determine whether differences in energy consumption between a variant and its corresponding baseline were statistically significant ($\alpha = 0.005$). Only variants meeting that criterion were included in the subsequent analysis.

%----------------------------------------------------------------------------------------------------
\section{Correctness of the Generated Code}\label{sec:correctness}
 
In this section we show how the use of contrastive prompt tuning impacts the correctness of the generated code. 

\vspace{0.2cm}
\noindent
\textbf{Baselines.} We start out by reporting baseline measurements for the three investigated language models. 
%This section presents the results of this work. Section~\ref{sec:res_baseline} lists the baseline accuracy results for the three investigated models. Section~\ref{subsec:res_exp_3} presents the results of an initial experiment examining the results for three contrastive loss functions. Based on the results of the initial experiment, Section~\ref{subsec:res_exp_4} reports on a broader set of experiments where we compare the results for the three languages, Java, C++, and Python, three models, using three datasets for CPT, and using the two best-performing functions, according to the preliminary experiment. Section~\ref{subsec:res_exp_5} compares the results obtained through CPT vs. giving an explicit instruction to try to improve performance in the prompt for the baseline models. Finally, Section~\ref{sec:code_energy_efficiency} reports on the energy efficiency of the programs generated with and without CPT. 
They serve as a reference for evaluating the prompt-tuned models. The corresponding results are presented in Table~\ref{tab:correctness}, in the rows labeled \texttt{baseline}. 
The general-purpose model has the highest total accuracy with 40.1\% correct solutions of a total of 648 tasks, with the code-tuned model close behind at 38.4\%. The code-first model achieves only 21.6\%. This difference appears across all benchmark sets, where the code-first model produces fewer correct solutions than the other models.  

Notably, the code-tuned model yields zero correct solutions on the HumanEval-X Java subset, despite leading on HumanEval-X C\texttt{++}, HumanEval-X Python, LeetCode C\texttt{++}, and LeetCode Python. The issue stems from the model generating repeated class definitions with the same name, which prevents Java files from compiling. A possible solution would be to process the result so that it also removes duplicate class declarations. 
%However, the pre-processing pipeline was kept unchanged, with the main reason being consistency. Earlier experiments used the same pre-processing pipeline and it was not feasible to re-run all previous experiments. Furthermore, 
Notwithstanding, for consistency, we have chosen not to do so. This behavior can be interpreted as an error of the model and we have not fixed incorrect programs for the other languages or models. Later experiments showed that variants of the code-tuned model did learn to generate compilable Java solutions for HumanEval-X.

Models are sensitive to how they are prompted. Therefore, we have also computed the accuracy of the models while adding a specific instruction about efficiency to the prompt, similarly to what we did in the initialization of the soft prompt during prompt tuning (Section~\ref{subsec:code_generation}). The instruction was \textit{``Write efficient code for the following problem''} and it was not part of the prompt used for the baseline models. We refer to this variant as the ``instruction'' variant and results for it are shown in Table~\ref{tab:correctness} in the rows labeled \texttt{instruction}.

%---------------------------------------------------------------------------

\vspace{0.2cm}
\noindent
\textbf{Models, languages, and training datasets.}
%In this section we present the results for the accuracy of the models variants prompt-tuned using the two loss functions and the three training sets. 
Table~\ref{tab:correctness} presents the accuracy of each model variant considering the two contrastive loss functions (``\texttt{infonce}'' and ``\texttt{triplet}'') used during prompt tuning for each training dataset (``\texttt{python}'', ``\texttt{cpp}'', ``\texttt{both}''), the two benchmark datasets (HumanEval-X and LeetCode), and the three languages in which code was generated. Results are reported for the epoch with the highest accuracy during training. 

%\subsubsection{Model: code-first}
% FIRST
\begin{table}[tb]
    \centering
    \renewcommand{\arraystretch}{1.2}
    \scriptsize
    \begin{tabularx}{0.48\textwidth}{|X|C|CCC|CCC|}
        \hline 
        & \multicolumn{7}{H|}{\textbf{Accuracy}}  \\
        \cline{2-8}
        \multicolumn{1}{|H|}{\textbf{Variant}} &  \multicolumn{1}{H|}{} & \multicolumn{3}{H|}{\textbf{HumanEval-X}} & \multicolumn{3}{H|}{\textbf{LeetCode}} \\
        \cline{3-8}
        \multicolumn{1}{|H|}{} &  \multicolumn{1}{H|}{\textbf{Total}} 
            &  \multicolumn{1}{H|}{\texttt{C++}} &  \multicolumn{1}{H|}{\texttt{Java}} & \multicolumn{1}{H|}{\texttt{Python}} 
            &  \multicolumn{1}{H|}{\texttt{C++}} &  \multicolumn{1}{H|}{\texttt{Java}} &  \multicolumn{1}{H|}{\texttt{Python}} \\
        \hline \hline
        \multicolumn{8}{|c|}{\textbf{Model: First}}\\ 
        \hline
                \texttt{baseline} 
            & 0.216 & 0.293 & 0.171 & 0.311 & 0.096 & 0.096 & 0.058 \\
        \cline{1-1}
                \texttt{instruction}  
            & 0.231 & 0.280 & 0.305 & 0.226 & 0.115 & 0.135 & 0.077 \\
        \hline
        \texttt{both-infonce-ep:1} 
            & 0.000 & 0.000 & 0.000 & 0.000 & 0.000 & 0.000 & 0.000 \\
        \cline{1-1}
        \texttt{cpp-infonce-ep:1} 
            & 0.000 & 0.000 & 0.000 & 0.000 & 0.000 & 0.000 & 0.000 \\      
        \cline{1-1}
        \texttt{python-infonce-ep:1} 
            & 0.171 & 0.268 & 0.140 & 0.201 & 0.077 & 0.077 & 0.058 \\
        \hline
        \texttt{both-triplet-ep:1}
            & 0.208 & 0.256 & 0.256 & 0.232 & 0.115 & 0.077 & 0.058 \\
        \cline{1-1}
        \texttt{cpp-triplet-ep:1} 
            & 0.207 & 0.280 & 0.207 & 0.250 & 0.096 & 0.096 & 0.058 \\
        \cline{1-1}
        \textcolor{darkgreen}{\texttt{python-triplet-ep:15}} 
            & 0.213 & 0.262 & 0.256 & 0.232 & 0.096 & 0.135 & 0.058 \\ 
          \hline\hline
        \multicolumn{8}{|c|}{\textbf{Model: General}}\\ 
        \hline 
        \texttt{baseline} & 0.401 & 0.378 & 0.457 & 0.543 & 0.173 & 0.154 & 0.327 \\
        \cline{1-1}
            \texttt{instruction} 
            & 0.457 & 0.421 & 0.549 & 0.610 & 0.154 & 0.173 & 0.385 \\
        \hline
        \texttt{both-infonce-ep:1} 
            & 0.465 & 0.433 & 0.561 & 0.616 & 0.154 & 0.154 & 0.404 \\
        \cline{1-1}
        \texttt{cpp-infonce-ep:1} 
            & 0.468 & 0.433 & 0.567 & 0.622 & 0.154 & 0.154 & 0.404 \\        
        \cline{1-1}
        \texttt{python-infonce-ep:1} 
            & 0.471 & 0.445 & 0.567 & 0.616 & 0.154 & 0.173 & 0.404 \\
        \hline
        \textcolor{darkgreen}{\texttt{both-triplet-ep:1}}
            & 0.474 & 0.445 & 0.567 & 0.616 & 0.173 & 0.173 & 0.423 \\
        \cline{1-1}
        \texttt{cpp-triplet-ep:1} 
            & 0.471 & 0.439 & 0.573 & 0.616 & 0.173 & 0.154 & 0.404 \\
        \cline{1-1}
        \texttt{python-triplet-ep:20} 
            & 0.471 & 0.433 & 0.585 & 0.610 & 0.173 & 0.154 & 0.404 \\
        \hline\hline
        \multicolumn{8}{|c|}{\textbf{Model: Tuned}}\\ 
        \hline 
        \texttt{baseline} & 0.384 & 0.628 & 0.000 & 0.659 & 0.212 & 0.135 & 0.385 \\
        \cline{1-1}
        \texttt{instruction}  
            & 0.432 & 0.646 & 0.299 & 0.512 & 0.250 & 0.173 & 0.365 \\
            \hline 
        \texttt{both-infonce-ep:1} 
            & 0.451 & 0.634 & 0.299 & 0.616 & 0.212 & 0.173 & 0.346 \\
        \cline{1-1}
        \texttt{cpp-infonce-ep:1} 
            & 0.441 & 0.640 & 0.329 & 0.537 & 0.231 & 0.173 & 0.346 \\  
        \cline{1-1}
        \texttt{python-infonce-ep:1} 
            & 0.448 & 0.616 & 0.335 & 0.585 & 0.231 & 0.173 & 0.327 \\  
        \hline
        \texttt{both-triplet-ep:20}  
            & 0.492 & 0.591 & 0.604 & 0.554 & 0.212 & 0.115 & 0.288 \\
        \cline{1-1}
        \textcolor{darkgreen}{\texttt{cpp-triplet-ep:20}}
            & 0.520 & 0.622 & 0.555 & 0.640 & 0.250 & 0.173 & 0.327 \\
        \cline{1-1}
        \texttt{python-triplet-ep:20} 
            & 0.461 & 0.634 & 0.366 & 0.604 & 0.212 & 0.173 & 0.308 \\
        \hline
    \end{tabularx}
    \caption{Pass@1 results (in \%) for all variants of the three models. Percentages are reported for the 648 tasks, followed by results on HumanEval-X and LeetCode, respectively 164 and 52 tasks per language: C\texttt{++}, Java, Python. The best-performing non-baseline, non-instruction variant for each model is indicated in \textcolor{darkgreen}{green} and the results pertain to the epoch with the best training results (``\texttt{ep}'') We report the baseline model accuracy for the original version, with no CPT.} 
    \label{tab:correctness}
\vspace*{-0.5cm}
\end{table}

For the code-first model, all the models, both baseline and prompt-tuned, exhibited poor performance. Variants prompt-tuned using the InfoNCE loss function performed particularly poorly, with two variants producing zero correct solutions. Overall, results were bad across training datasets and benchmark datasets, although worse for the LeetCode dataset, e.g., the highest accuracy for LeetCode was 13.5\%.
% for the variant trained on the Python dataset using Triplet loss. 
The mean accuracy was 21.6\% for the baseline, 23.1\% for the instruction variant (a 1.5 percentage-point gain), and 21.3\% for the best-performing prompt-tuned model, using the Triplet loss function with the Python training set at epoch 15. 
The best prompt-tuned model only surpassed the baseline for Java code generation, with 8.5\% and 3.9\% improvements in accuracy for HumanEval-X and LeetCode, respectively. It is interesting to note that the performance of the model variants improved for Java even though they were only trained on Python and C++ examples. 

%There is a clear performance gap between InfoNCE and Triplet loss. For InfoNCE, two variants yield zero correct solutions, and only the Python training set at epoch 1 yields 17.1\% correct solutions. Triplet variants achieve between 20.7\% and 21.3\% correct solutions, representing an improvement of 3.6 to 4.2\% over the best InfoNCE variant. This small difference is mostly due to higher accuracies on the HumanEval-X Java and Python subsets.  

%Comparing the performance between training sets shows that both losses perform best with the Python training set: InfoNCE yields 17.1\% and Triplet yields 21.3\%. InfoNCE with C\texttt{++} and combined training sets (Both) performs very poorly, whereas Triplet results remain stable across training sets. 
%When focusing on Triplet variants, the C\texttt{++} training set yields fewer correct Java solutions by 4.9\% compared to Python and combined sets, but yields slightly higher C\texttt{++} and Python solutions by 2.5 and 1.8\%. Thus, in our experiments the choice of training set strongly affected InfoNCE but had only minor impact for Triplet. 

%As with the general-purpose model, Triplet with minimal training (epoch 1) already shows benefit: the Python Triplet variant at epoch 1 achieves 20.7\%, only 0.6\% lower than epoch 15. These findings show that \Gls{prompttuning} for the code-first model yields effects after minimal training. Nevertheless, accuracy remains below the baseline, suggesting limited potential for further improvement with \gls{cpt}. \\

For the general-purpose model, Table~\ref{tab:correctness} shows that every combination using InfoNCE or Triplet loss improved accuracy compared to the baseline. The baseline achieved 40.1\% correct solutions and the instruction variant 45.5\%. Notwithstanding, the accuracy of the prompt-tuned variants ranged from 46.5\% to 47.4\%, with the highest mean accuracy of 47.4\% achieved by the variant using the Python training set and Triplet loss at epoch 1.
% This is in stark contrast to the results obtained for the code-first model, where the best performing prompt-tuned variant showed slightly worse results than the corresponding baseline. 
Comparing InfoNCE and Triplet loss shows that Triplet performs slightly better, but the difference is minimal, with only 0.9\%. At the level of individual benchmarks, differences among loss functions and training sets are small.  
Examining the training sets reveals that the combined training set (Both) yields the best Triplet result at 47.4\%, while InfoNCE on the combined set (Both) reaches 46.5\%; this small gap indicates that training set choice had little impact for the general purpose model. %This contrasts with the findings for the code-first model. \\
Most variants reach peak performance early: five variants perform best at epoch 1, and the Python Triplet variant at epoch 1 achieves 47.4\% compared to 47.0\% at epoch 20, a difference of only 0.4\%, suggesting that it was possible to achieve \textbf{significant benefits from a small amount of contrastive prompt tuning}.  \\

As observed for the code-first model, the performance of the variants when generating Java code improved compared to the baseline in spite of Java examples not being part of the training set. Similarly, variants prompt-tuned only with Python examples improved their accuracy when generating C++ code. This indicates that \textbf{the learned representations generalize across languages}.

The code-tuned model and its variants also benefit from contrastive prompt tuning, as each variant yields higher mean accuracy than the mean baseline accuracy of 38.4\% and the mean instruction accuracy of 43.2\%. The best variant uses Triplet loss with the C\texttt{++} training set at epoch 20, achieving 52.0\% correct solutions. All variants improve by at least 5.7\% on mean accuracy over the baseline and 0.9\% on the instruction variant. A notable factor is that the baseline has zero correct solutions  and instruction has 29.9\% accuracy for HumanEval-X Java, which strongly influences the overall result. Examining benchmark-level performance shows that all variants underperform the baseline on HumanEval-X Python and LeetCode Python by at least 3.0 and 2.0\% respectively. This is unsurprising, as the baseline version was fine-tuned mostly on Python code\footnote{https://huggingface.co/codellama/CodeLlama-7b-Instruct-hf}. In contrast, multiple variants surpass the baseline, considering both datasets, in Java and C\texttt{++}. 

When compared against the instruction variant, all variants surpass it on HumanEval-X Python and Java, with one exception. In contrast, the instruction variant on HumanEval-X C\texttt{++} has the highest accuracy for that language. On LeetCode, the instruction variant exhibits the best results for C\texttt{++} and Java, 25\% and 17.3\%, respectively, a tie with the Triplet loss variant trained on the C\texttt{++} training set. 
%These observations confirm that \textbf{the missing HumanEval-X Java solutions in the baseline cause much of the overall improvement}. \\

Comparing InfoNCE and Triplet loss indicates that Triplet yields higher overall accuracy: Triplet variants range from 46.1\% to 52.0\% correct solutions, which is 1.0 to 6.9\% higher than the best InfoNCE variant at epoch 1 (45.1\%). This makes \textbf{Triplet the superior choice for code correctness} for this model. For InfoNCE, the best results occur at epoch 1 and decline with more epochs, whereas Triplet improves with longer prompt tuning. Since longer tuning also consumes more energy, InfoNCE may be preferable when energy cost is a priority but Triplet is better for accuracy.
Considering the three training sets for InfoNCE, overall accuracy differences are small except on specific subsets: on HumanEval-X Python, the combined train set (Both) outperforms the C\texttt{++} training set by about 7.9\%, while on HumanEval-X Java, the C\texttt{++} training set outperforms the combined training set (Both) by roughly 3.0\%. For Triplet loss, the choice of training set matters more: Tuning with the Python training set results in about 5.9\% lower overall accuracy than the C\texttt{++} set, mostly caused by an 18.9\% drop on HumanEval-X Java. The combined training set (Both) achieves the highest Java accuracy (~60.4\%), but is behind the C\texttt{++} training set on other cases by 3–9\%. This contrasts with the code-first model’s behavior and shows that \textbf{optimal training set selection varies by model variant}. \\
\section{Energy efficiency of the generated code}
\label{sec:code_energy_efficiency}

This section presents energy consumption measurements of the generated code, for baseline models and their variants (best, instruction).
%For each model, the baseline variant, the instruction variant, and the best variant identified in Section~\ref{subsec:res_exp_5} are evaluated. 
%In addition, an alternative variant is included for the general-purpose and code-tuned models to allow further comparison.
We report only on results that reach statistical significance compared to the baseline. Table~\ref{tab:summary_metrics} presents the results as the mean percentage of the energy and execution time consumed by programs generated by a baseline model. 
We only perform comparisons if both the baseline model and a variant (best or instruction) were able to successfully generate code. Therefore, the number of comparisons for best and instruction in the same row may differ. The use of mean percentages avoids big discrepancies that can arise in this scenario. Furthermore, we argue that the performance of an incorrect program does not matter. An alternative would be to report only on the performance of the common sub-set of correctly generated programs, but this would greatly reduce the amount of reported data. 
%The LeetCode benchmark (N = 52) is 68.29\% smaller than HumanEval-X (N = 164), which limits the ability to draw strong conclusions from it, especially regarding the effects of training on more complex tasks. Nevertheless, some general trends can be observed.

Across all three model variants, training tends to have a more consistent impact on energy efficiency and execution time than instruction. Differences between best-trained and instruction variants are often large. For example, the first model shows an energy efficiency of 58.5\% for HumanEval-X C\texttt{++} in the best variant versus 174.5\% with instruction. Figure~\ref{fig:code_example} presents examples of programs generated for a programming problem from HumanEval-X. The inefficient program on the left-hand side was generated by the baseline code-first model whereas the one on the right-hand side is a more efficient solution produced by the best prompt-tuned variant. The efficient version includes early exits for known cases and skips even numbers, reducing loop iterations by half.
%This demonstrates how minimal algorithmic refinements can yield notable efficiency improvements.
In addition, for LeetCode Java (code-first and code-tuned) and LeetCode C\texttt{++} (code-first) no program generated through the instruction approach exhibited performance that differed from the baseline with statistical significance. For the best variant, this happened in just one scenario, LeetCode Python (code-first). These results suggest that while \textbf{training could improve energy efficiency, the results are uncertain}, considering the  small sample size of this study. 

\lstdefinestyle{cppstyle}{
    language=C++,
    basicstyle=\ttfamily\scriptsize,
    keywordstyle=\color{blue},
    commentstyle=\color{gray},
    stringstyle=\color{red},
    breaklines=true,
    tabsize=2,
    frame=single,
    numbers=left,
    numberstyle=\tiny,
    numbersep=5pt
}
\begin{figure}[tb]
\scriptsize
\centering
\begin{tabular}{ll}
\begin{minipage}{0.22\textwidth}
\centering
\caption*{Inefficient code solution}
\begin{lstlisting}[style=cppstyle]
bool is_prime(long n){
  if(n == 1)
    return false;
  for(long i=2; i*i<=n; i++){
    if(n % i == 0)
      return false;
  }
  return true;
}
\end{lstlisting}

\end{minipage} & 
\hfill
\begin{minipage}{0.22\textwidth}
\centering
\caption*{Efficient code solution}
\begin{lstlisting}[style=cppstyle]
bool is_prime(long n){
  if(n == 1) return false;
  if(n == 2) return true;
  if(n % 2 == 0) return false;
  for(int i=3; i*i<=n; i+=2){
    if(n%i == 0) return false;
  }
  return true;
}
\end{lstlisting}
\end{minipage} 
\end{tabular}
\caption{Comparison of inefficient (left) and efficient (right) for HumanEval-X C\texttt{++} problem 31. }
\label{fig:code_example}
\end{figure}

CPT also yielded worse results in some scenarios. In LeetCode Java for the general-purpose model, training increases energy usage (207.6\%), and the instruction variant offers a clear improvement (92.0\%). A more nuanced example is LeetCode Python for the code-tuned  model, where the prompt-tuned model exhibits slightly higher energy consumption (104.4\%) while the instruction variant has slightly lower energy usage (96.6\%). 

Execution time largely mirrors energy consumption. When training yields energy improvements, it typically also reduces execution time. Furthermore, the changes tend to have the same magnitude, e.g., the prompt-tuned, code-tuned model on HumanEval-X for C\texttt{++} produces programs that on average consume 98.3\% of the energy and 98.5\% of the time of the baseline-generated programs.

\begingroup
  \setcellgapes{3pt}
  \makegapedcells
  \begin{table}[tb]
      \centering
      \renewcommand{\arraystretch}{1.2}
      \scriptsize
      \begin{tabular}{|l|l|l||c|c|c|c|}
          \hline
              & 
              &
%              & \multicolumn{2}{c|}{\textbf{Accuracy$\uparrow$}} 
              & \multicolumn{2}{c|}{\textbf{Energy$\downarrow$}} 
              & \multicolumn{2}{c|}{\textbf{Exec. Time$\downarrow$}} \\
          \cline{4-7}
          \textbf{Model} 
              & \textbf{Benchm.} 
              & \textbf{Trainset} 
%              & \textbf{best} 
%              & \textbf{instr.} 
              & \textbf{best} 
              & \textbf{instr.} 
              & \textbf{best} 
              & \textbf{instr.} \\
          \hline\hline
          \multirow[c]{12}{*}{first} 
              & \multirow[c]{6}{*}{H.Eval-X} 
              & C\texttt{++}     % & 89.4\%   & 95.6\%   
              & \makecell[c]{58.5\%}  & \makecell[c]{174.5\%} & \makecell[c]{57.9\%}  & \makecell[c]{172.4\%} \\
          \cline{3-7}
              &                               
              & Java    % & 149.7\%  & 178.4\%  
              & \makecell[c]{153.0\%} & \makecell[c]{45.3\%}  & \makecell[c]{154.7\%} & \makecell[c]{45.7\%}  \\
          \cline{3-7}
              &                               
              & Python   %& 74.6\%   & 72.7\%  
               & \makecell[c]{86.2\%}  & \makecell[c]{87.4\%}  & \makecell[c]{88.7\%}  & \makecell[c]{92.2\%}  \\
          \cline{2-7}
              & \multirow[c]{6}{*}{LeetCode} 
              & C\texttt{++}    %  & 100.0\%  & 119.8\% 
               & \makecell[c]{33.2\%}  &                            & \makecell[c]{34.5\%}  &                            \\
          \cline{3-7}
              &                               
              & Java   %  & 140.6\%  & 140.6\% 
               & \makecell[c]{163.1\%} &                            & \makecell[c]{165.6\%} &                            \\
          \cline{3-7}
              &                               
              & Python %  & 100.0\%  & 132.8\%  
              &                            & \makecell[c]{94.7\%}  &                            & \makecell[c]{95.9\%}  \\
          \cline{2-7}
          \hline
          \multirow[c]{12}{*}{general} 
              & \multirow[c]{6}{*}{H.Eval-X} 
              & C\texttt{++}    %  & 117.7\%  & 111.4\%  
              & \makecell[c]{83.8\%}  & \makecell[c]{90.7\%}  & \makecell[c]{84.8\%}  & \makecell[c]{90.8\%}  \\
          \cline{3-7}
              &                               
              & Java   %  & 124.1\%  & 120.1\%  
              & \makecell[c]{228.4\%}& \makecell[c]{128.9\%} & \makecell[c]{235.9\%}& \makecell[c]{120.9\%} \\
          \cline{3-7}
              &                                
              & Python %  & 113.4\%  & 112.3\%  
              & \makecell[c]{111.0\%}& \makecell[c]{129.1\%}& \makecell[c]{111.4\%}& \makecell[c]{129.7\%}\\
          \cline{2-7}
              & \multirow[c]{6}{*}{LeetCode} 
              & C\texttt{++}     % & 100.0\%  & 89.0\%   
              & \makecell[c]{98.4\%}  & \makecell[c]{98.3\%}  & \makecell[c]{98.9\%}  & \makecell[c]{98.5\%}  \\
          \cline{3-7}
              &                               
              & Java    % & 112.3\%  & 112.3\%  
              & \makecell[c]{207.6\%} & \makecell[c]{92.0\%}  & \makecell[c]{200.8\%} & \makecell[c]{92.2\%}  \\
          \cline{3-7}
              &                               
              & Python  % & 129.4\%  & 117.7\% 
               & \makecell[c]{83.9\%}  & \makecell[c]{95.0\%} & \makecell[c]{82.0\%}  & \makecell[c]{95.1\%} \\
          \cline{2-7}
          \hline
          \multirow[c]{10}{*}{tuned} 
              & \multirow[c]{4}{*}{H.Eval-X} 
              & C\texttt{++}    %  & 99.0\%   & 102.9\% 
               & \makecell[c]{119.5\%}& \makecell[c]{107.7\%} & \makecell[c]{119.6\%}& \makecell[c]{109.0\%} \\
          \cline{3-7}
              &                               
              & Python %  & 97.1\%   & 77.7\%  
               & \makecell[c]{92.9\%} & \makecell[c]{86.7\%} & \makecell[c]{91.8\%} & \makecell[c]{90.7\%}\\
          \cline{2-7}
              & \multirow[c]{6}{*}{LeetCode} 
              & C\texttt{++}    %  & 117.9\%  & 117.9\% 
               & \makecell[c]{104.8\%} & \makecell[c]{107.6\%} & \makecell[c]{102.6\%} & \makecell[c]{107.1\%} \\
          \cline{3-7}
              &                               
              & Java   %  & 128.1\%  & 128.1\% 
               & \makecell[c]{44.0\%}  &                            & \makecell[c]{47.6\%}  &                            \\
          \cline{3-7}
              &                               
              & Python  % & 84.9\%   & 94.8\%  
               & \makecell[c]{104.4\%} & \makecell[c]{96.6\%}  & \makecell[c]{103.2\%} & \makecell[c]{97.4\%}  \\
          \cline{2-7}
          \hline
      \end{tabular}
      \caption{Percentage values for best-performing trained and instruction variants of each model, benchmark, and training set, relative to the baseline (baseline = 100\%), for energy consumption and execution time. 
      %Numbers in parentheses show the count \(N\) of tasks used in each comparison (only cases with statistically significant difference are counted). 
      A percentage lower than 100\% is a positive result. Blank cells and missing rows indicate non-statistically-significant or unavailable data.}
      \label{tab:summary_metrics}
      \vspace*{-0.3cm}
  \end{table}
\endgroup

%-----------------------------------------------------------
\section{Discussion}\label{sec:discussion}

\noindent
\textbf{CPT Improves Accuracy.}
Prompt tuning with %text-based initialization and 
contrastive learning improved the accuracy for the general-purpose and code-tuned models across benchmarks, most notably for HumanEval-X. This aligns with the findings by Jain et al.~\cite{jain_contraclm_2023}, who also reports accuracy gains on the HumanEval benchmark. In contrast, this work extends those insights by evaluating the models on HumanEval-X, demonstrating that similar improvements can be achieved for Java and C\texttt{++} as well. Therefore, it seems that training with pairs of correct solutions that belong to the same problem but differ in efficiency, also helps the model to learn patterns of correct code.  
%Although the primary aim of this work is to only preserve correctness while improving energy efficiency, higher accuracy also reduces debugging effort and thus saves energy and time downstream. 

\vspace{0.1cm}
\noindent
\textbf{Task and Cross-Language Generalization.}
Many of the largest accuracy/efficiency improvements occur on HumanEval-X, which consists of relatively straightforward coding tasks. Improvements on LeetCode are smaller or mixed, reflecting the greater difficulty and diversity of those problems. Java subsets show consistent gains across models after CPT, despite the absence of Java examples in training. Training solely on Python data also led to improvements on both C\texttt{++} and Java benchmarks. These cross-language effects suggest that soft prompts and contrastive learning guide the model towards solution strategies that go beyond syntactic differences. CPT seems to help the model account for underlying algorithmic structures and correctness conditions, thereby supporting the conclusion that CPT captures functional semantics in a language-agnostic way. Furthermore, similar cross-language generalization effects were observed by Li et al.~\cite{li_coderetriever_2022}, who demonstrated that contrastive learning improved their model’s ability to align code representations across programming languages in a code search setting: they fine-tuned on Python query-code pairs and evaluated on Java.

\vspace{0.1cm}
\noindent
\textbf{Model Architecture Matters.}
The three model variants display different learning dynamics under prompt CPT, reflecting their pre-training and architectural backgrounds. The code-first model (DeepSeek Coder, ~7B parameters), which was designed from the ground up for code generation, shows minimal or negative gains from CPT in code correctness and code energy-efficiency. One reason could be that its representations are already closely aligned with code-like inputs and outputs, so the approach of this work results in limited additional benefit. At the same time, the data suggests that, even if this hypothesis is correct, these learned representations do not lead to improved results. Baseline accuracy for the general-purpose model was 40.1\%, compared to the 21.6\% for the code-first model. Even then, the best-performing prompt-tuned variant of the former still exhibited an improvement of 7.3\%. 

The general-purpose model (Phi-3 Mini, 3.8B parameters) benefits from just one epoch of contrastive tuning. The embedding space seems to adapt quicker to distinguishing correct from incorrect code patterns, which lead to significant accuracy improvements with minimal training. In comparison, the code-tuned model (CodeLlama-7B, 7B parameters), pre-trained on large code corpora, requires longer prompt-tuning cycles. This could stem from the model's representations already encoding code structures well, but they requiring extended contrastive training to capture fine-grained constraints like energy-efficient patterns. 
%These are novel insights, as none of the related studies evaluated their methods across different model types. 
%The results suggest that models that code-first models may have less room for CPT gains, while general-purpose models can quickly benefit by specializing their embeddings. Code-tuned models, though already familiar with code, still demand more extensive training to align their embeddings. In practice, this underlines the importance of customizing tuning schedules. However, this work only tested one model per architecture, so conclusions drawn are not generalizable. 

\vspace{0.1cm}
\noindent
\textbf{Prompting and CPT}.  
The results show that adding an efficiency instruction to the prompt can improve accuracy across all models with no extra training energy costs. For the code-first model, the instruction is even the only beneficial change in terms of accuracy. These findings resonate with those of Cappendijk et al.~\cite{cappendijk_generating_2024} and Apsan et al.~\cite{Apsan:2026:GEE}, who showed that well-crafted efficiency instructions can positively influence energy consumption of LLM-generated code. Although their results were mixed across different models and tasks, they underscore the potential of simple prompt modifications. However, Waghjale et al.~\cite{waghjale_ecco_2024}  reported that their instruction prompting decreased accuracy but improved energy efficiency. However, their approach involved more complex instructions, which included examples of both efficient and inefficient code. This differs significantly from the simpler instruction and more sophisticated CPT approach used in this work. Nevertheless, this does show how the design of prompt instructions can significantly influence both accuracy and energy outcomes.  

For the general-purpose model, one epoch of contrastive prompt-tuning on top of the instruction is enough for further gains. For the code-tuned model, extended tuning results in additional improvement, but the initial instruction already boosts accuracy significantly, especially for Java correctness. Thus, we can say that, at least for this initial exploration, our results were mixed. 
\section{Threats to Validity}\label{section:threats}

%This section addresses multiple threats to validity that might impact the reliability or generalizability of the results. The threats are addressed in the order in which they would appear during the stages of this work.  
\noindent
\textbf{Scope and generalizability.} 
The chosen models might not reflect the current landscape of LLMs properly and therefore conclusions might not be generalizable. To mitigate this threat, three models with different underlying architectures were chosen. This does not address the problem that there are many different models available, but it is not feasible to cover a wide range of models and also explore variations along other dimensions, such as programming languages, contrastive loss functions, and training datasets. Another threat is posed by the choice of programming languages that were evaluated, as different languages exhibit different characteristics. Examining only one programming language would therefore limit the generalizability of the results. This work tries to reduce that threat by evaluating Python, Java and C\texttt{++}, since previous work ~\cite{solovyeva_ai-powered_2025} has shown that these three languages perform very differently when it comes to LLM-generated code. Finally, we have not evaluated the impact of different prompting strategies on the efficiency of the generated programs, as it would have required a completely different study. We only measured the impact of including an instruction about efficiency on  accuracy, with and without CPT.

\vspace{0.1cm}
\noindent
\textbf{CPT training datasets.}  The selection of the training datasets for CPT was based on the assumption that efficient programs, as established by the developers of the datasets, consistently outperform their less efficient, functionally equivalent variants in execution time and energy consumption. Thus, the loss function does not need to directly account for the performance or energy efficiency of the generated programs, only their correctness. One the one hand, performance is platform-dependent and explicitly accounting for it could have led to improved results. On the other hand, we experimented with only one machine and results would arguably not be transferrable to other settings. Future work can explore the use of performance measurements (or estimates) during CPT, as they can directly impact the distance between representations in the embedding space, across different platforms.

\vspace{0.1cm}
\noindent
\textbf{Implementation Choices and Dataset Representativeness.}
Since variations in the implementation of CPT can have a great effect on the outcome, an attempt was made to use standard, well-known implementations. Therefore, the models are based on the HuggingFace Transformer library and the prompt tuning implementation on HuggingFace’s PEFT library. Furthermore, as in any fine-tuning scenario, the used dataset determines the success heavily. As this work requires the construction of a new dataset, a risk is introduced that it might be incomplete or non-representative. To mitigate these risks, the publicly available datasets GEC and CodeNet were used as a basis. Furthermore, the choice of benchmarks also poses a threat, as a benchmark might not capture the complexity of real-world scenarios effectively. %Many benchmarks currently used to evaluate the generated code of LLM, mostly include easy programming tasks. 
To mitigate this threat, we also used a benchmark~\cite{solovyeva_ai-powered_2025} consisting of hard coding problems from LeetCode, together with a well-established benchmark, HumanEval-X, consisting of easy problems.% By doing so, a more realistic scenario with varying difficulty levels was investigated. %Lastly, in regards to the code generation phase, some code also required pre-processing before being evaluated. To ensure consistency in this process, the same augmentations that were pre-defined were applied to each code solution. 

\vspace{0.1cm}
\noindent
\textbf{Measurement Procedures and Evaluation.}
The stage of evaluating the code and drawing conclusions also holds some threats. Only one machine (MacBook, 2016) was used for measurements. Here, the hardware and software specifics of the machine might limit the generalizability of the results. Furthermore, the specific settings together with background processes might influence the outcome of the measurements. To mitigate these threats, the evaluations were performed under the same controlled conditions, as described in Section~\ref{sec:eval}. Additionally, each problem case was run 10 times to reduce the risk of variability in the results due to thermal throttling and (in the case of Java) the JVM not being hot~\cite{Pinto:2014:UEB}. 
%Lastly, the way the energy-efficiency is determined greatly influences the results. Only depending on one metric might mean a lower chance of generalizability and reliability. To mitigate this risk, in addition to the code correctness, the metric of execution time and energy consumption was used to evaluate the energy-efficiency of the generated code. 

%----------------------------------------------------------------------------------------------------
\section{Concluding Remarks}

In this paper we have shown that CPT may be a path to generate energy-efficient programs. 
The main insight of this work lies in leveraging the discriminating ability of contrastive approaches~\cite{Chen:2020:SFC} to learn the differences between efficient and inefficient implementations of functionally equivalent To the best of our knowledge, this is the first paper to explore the potential of combining contrastive learning with prompt tuning to generate programs with improved runtime characteristics, i.e., lower energy footprint. 
The use of CPT lead to models more reliably producing correct code. Moreover,
in some cases, it was possible to achieve expressive (up to 66\% for C\texttt{++}, up to 56\% for Java) reductions in energy usage compared to baseline results produced by the models. The results were mixed, though, with energy use increasing in a similar number of cases.

%%%%%%%%%%%%%%%%%%%%%%%%%%%%%%%%%%%%%%%%%%%%%%%%%
\bibliographystyle{ACM-Reference-Format}
\bibliography{llm4code_2026}

\end{document}